\documentclass[a4paper, 10pt, conference]{ieeeconf}      % Use this line for a4
\IEEEoverridecommandlockouts                              % This command is only
\makeatletter
\let\NAT@parse\undefined
\makeatother

\usepackage[dvipsnames]{xcolor}

\newcommand*\linkcolours{ForestGreen}

\usepackage{times}
\usepackage{graphicx}
\usepackage{amssymb}
\usepackage{gensymb}
\usepackage{mathtools}
\usepackage{breakurl}
\usepackage{blindtext}

\usepackage{url,hyperref}
\hypersetup{
colorlinks,
linkcolor=\linkcolours,
citecolor=\linkcolours,
filecolor=\linkcolours,
urlcolor=\linkcolours}

\usepackage{algorithm}
\usepackage{algorithmic}

\usepackage[labelfont={bf},font=small]{caption}
\usepackage[none]{hyphenat}

\usepackage{mathtools, cuted}

\usepackage[noadjust, nobreak]{cite}
\usepackage{tabularx}
\usepackage{amsmath}

\usepackage{float}

\usepackage{pifont}% http://ctan.org/pkg/pifont

\newcolumntype{Y}{>{\centering\arraybackslash}X}

\usepackage[]{placeins}

\newcommand\extraspace{3pt}

\usepackage{placeins}

\usepackage{tikz}

\usepackage[framemethod=tikz]{mdframed}

\usepackage{afterpage}

\usepackage{stfloats}

\usepackage{atbegshi}
\newcommand{\handlethispage}{}
\newcommand{\discardpagesfromhere}{\let\handlethispage\AtBeginShipoutDiscard}
\newcommand{\keeppagesfromhere}{\let\handlethispage\relax}
\AtBeginShipout{\handlethispage}

\usepackage{comment}

\title{\LARGE \bf
Cross-lingual paraphrase identification
}

\author{Inessa Fedorova$^{1}$ and Aleksei Musatow$^{2}$% <-this % stops a space
\thanks{$^{1}$AI Talent Hub, 
ITMO University, Saint-Petersburg
Email: 412549@edu.itmo.ru}%
\thanks{$^{2}$SaluteDevices, 
Saint-Petersburg,
Email: musattow@gmail.com}%
}

\begin{document}

\maketitle
\thispagestyle{empty}
\pagestyle{empty}

%%%%%%%%%%%%%%%%%%%%%%%%%%%%%%%%%%%%%%%%%%%%%%%%%%%%%%%%%%%%%%%%%%%%%%%%%%%%%%%%
\begin{abstract}
The paraphrase identification task involves measuring semantic similarity between two short sentences. It is a tricky task, and multilingual paraphrase identification is even more challenging.

In this work, we train a bi-encoder model in a contrastive manner to detect hard paraphrases across multiple languages. This approach allows us to use model-produced embeddings for various tasks, such as semantic search.

We evaluate our model on downstream tasks and also assess embedding space quality. Our performance is comparable to state-of-the-art cross-encoders, with only a minimal relative drop of 7-10\% on the chosen dataset, while keeping decent quality of embeddings.

\end{abstract}

%%%%%%%%%%%%%%%%%%%%%%%%%%%%%%%%%%%%%%%%%%%%%%%%%%%%%%%%%%%%%%%%%%%%%%%%%%%%%%%%
\section{Introduction}

Paraphrase identification (PI) is a fundamental task in natural language processing (NLP), where the goal is to determine whether a pair of sentences express the same or similar meaning 1. Researchers have proposed various methods for PI, which find applications in diverse areas, including:
\begin{itemize}
\item Translation plagiarism detection
\item NMT evaluation
\item Parallel corpora creation
\end{itemize}
\vspace{\extraspace}

During development, automatic measures are commonly used to assess sentence similarity. These measures compute a value (denoted as sim(x, y)) indicating the similarity between an original text (x) and a paraphrased text (y), typically within the range [0, 1]. Higher scores indicate greater content similarity, while lower scores indicate differences in content.

In this work, we demonstrate that bi-encoders can be effectively applied to the paraphrase identification task. We train a multilingual bi-encoder model using contrastive loss, aiming to achieve quality comparable to Cross-encoders models. Additionally, we employ specific training techniques to enhance the model’s performance.

\vspace{\extraspace}
Thus, the contribution of our work can be summarized as follows:
\begin{itemize}
\item We propose an approach for training bi-encoders for solving paraphrase identification task
\item We modify Additive Margin Softmax Loss and sample hard-negatives to make the model's training process more effective
\item We evaluate the model on the PAWS-X dataset and use embeddings space metrics, comparable to state-of-the-art cross-encoders a minimal relative drop of 7-10\%.
\end{itemize}

%%%%%%%%%%%%%%%%%%%%%%%%%%%%%%%%%%%%%%%%%%%%%%%%%%%%%%%%%%%%%%%%%%%%%%%%%%%%%%%%
\section{Related work}
\subsection{Paraphrase identification techniques}
In this section, we explore existing approaches for measuring semantic similarity.

\textbf{N-gram-based metrics}: Comparing two texts based on the overlap of word or character n-grams is intuitive. Simple distance-based approaches like Levenshtein distance can be used, but more effective methods include standard metrics such as BLEU score \cite{papineni2002bleu}, ROUGE \cite{lin2004rouge}, and METEOR \cite{banerjee-lavie-2005-meteor}. However, n-gram measures do not handle linguistic phenomena like synonyms, negation, and word order issues effectively.

\textbf{Similarity between static embeddings}: Another family of measures partially overcomes the aforementioned difficulties. It relies on calculating distances between text embeddings. These measures can be further categorized based on how the embeddings are generated. The basic approach involves averaging static word embeddings, such as Word2vec \cite{mikolov2013efficient}, GloVe \cite{pennington2014}, and FastText \cite{bojanowski2017enriching}.

\textbf{Similarity between embeddings from bi-encoders}:  Text embeddings can be generated by encoding a text with a pre-trained encoder. If the two texts are encoded separately, and we compute the cosine similarity between their embeddings, we refer to these models as bi-encoders. Typically trained in a supervised manner, these encoders can be fine-tuned for tasks like translation (Laser \cite{Artetxe_2019}, LaBSE\cite{feng2022languageagnostic}), paraphrase identification (SIMILE \cite{wieting2019bleu}), or text generation (BARTScore \cite{yuan2021bartscore}). They have the potential to compare text meanings while preserving context. 

\textbf{Cross-encoders}:
Cross-encoders process both texts simultaneously using cross-attention and directly predict the relationship between the texts. They can perform symmetrically (where the score is independent of the order of the texts being compared) or asymmetrically (where the score strongly depends on the order of the texts). Due to their supervised nature, such models can reflect content preservation more accurately than word-based approaches. However, they depend on labeled data and may not generalize well to new domains.

The presence of symmetry is defined by the task the model was trained on. For instance:

\begin{itemize}
\item Models trained on Natural Language Inference (NLI) task data (such as BLEURT \cite{sellam2020bleurt} or NUBIA \cite{kane-etal-2020-nubia}) are asymmetric.
\item Cross-stsbbase models trained solely on the STS-B dataset \cite{Cer_2017} for semantic textual similarity or APD model \cite{nighojkar-licato-2021-improving} trained on paraphrase datasets perform symmetrically.
\end{itemize}
\vspace{\extraspace}
Based on the information listed above we decided to use bi-encoders, because of such their advantages:
\begin{itemize}
\item Potentially can compare the meanings of texts that are very different in terms of structure and vocabulary
\item Ability to derive semantically meaningful sentence embeddings
\item Computational efficiency (lower complexity, offline sentence encoding)
\item Easier new domain generalization
\end{itemize}
\vspace{\extraspace}

\subsection{Contrastive learning}

Contrastive learning aims to learn effective representation by pulling semantically close neighbors
together and pushing apart non-neighbors \cite{handsell2006}. It assumes a set of paired examples
where each pair are semantically related.

We follow the idea of \cite{yang2019improving} and take a Additive Margin Softmax objective with in-batch negatives. Additive Margin Softmax extends the scoring function by the large margin which improves the separation between positive and negative examples.

\textbf{Positive and negative instances} One critical question in contrastive learning is how to construct positive pairs.
In visual representations, an effective solution is to take two random transformations of the same image
(e.g., cropping, flipping, distortion and rotation) as positive pair \cite{dosovitskiy2015discriminative}. A similar approach had been adopted in language representations (\cite{wu2020clear}; \cite{meng2021cocolm}) by applying augmentation techniques such as word deletion, reordering, and substitution. However, data augmentation in NLP is inherently difficult because of its discrete nature. As was shown in \cite{gao2022simcse} simply using standard dropout on intermediate representations outperforms these discrete operators.

In NLP, a similar contrastive learning objective has been explored in different contexts (\cite{henderson2017efficient}; \cite{gillick-etal-2019-learning}; \cite{karpukhin2020dense}). In these cases, positive pairs are collected from supervised datasets such as question-passage pairs and negative pairs selected from the rest of batch (in-batch negatives) or also collected from supervised dataset. There are also other effective techniques to create more complex examples pairs which would be discussed in experiments section. 

\vspace{\extraspace}
%%%%%%%%%%%%%%%%%%%%%%%%%%%%%%%%%%%%%%%%%%%%%%%%%%%%%%%%%%%%%%%%%%%%%%%%%%%%%%%%
\section{Experiments setup}

\subsection{Dataset and sampling details}

\textbf{Dataset description} We train and evaluate the model on the PAWS-X dataset\cite{yang2019pawsx}.  The training portion of the dataset consists of 49,401 machine-translated paraphrases. The test and development sets are human-translated, each containing 2,000 paraphrases.

Paraphrases for each text were created using the following methods:

\begin{itemize}
\item Adjective swap
\item Named entity swap
\item Verb swap
\item Word replacement
\end{itemize}

PAWS-X includes paraphrases in 7 languages. Based on language criteria, we can divide the paraphrases in the dataset into 3 types:
\begin{itemize}
\item Intra-lingual: When we compare phrases with similar meanings in the same language.
\item Inter-lingual translation: When we compare a phrase in one language to its translation in another.
\item Inter-lingual paraphrase: When we compare a phrase in one language with a phrase having a similar meaning in another language.

\end{itemize}

\textbf{Sampling details}

Sampling, as the process of constructing pairs, is one of the most challenging aspects of contrastive learning. Due to the dataset content, we have decided to use three classes of phrase pairs:
\begin{itemize}
\item \textbf{Positive pairs (positives)}: These are pairs of sentences from the dataset with a class label equal to 1 (indicating similarity).
\item \textbf{In-batch negative pairs (negatives)}: These pairs consist of sentences selected from the same batch but not related to each other. Since this combination does not exist in the dataset, we treat it as a pair with phrases having dissimilar meanings.
\item \textbf{Hard negative pairs (hard negatives)}: These pairs consist of sentences from the dataset with a class label equal to 0 (indicating dissimilarity).
\end{itemize}
Examples of paraphrases and their relationships are shown in Figure 1.

\begin{figure}
    \centering
    \includegraphics[width=1\linewidth]{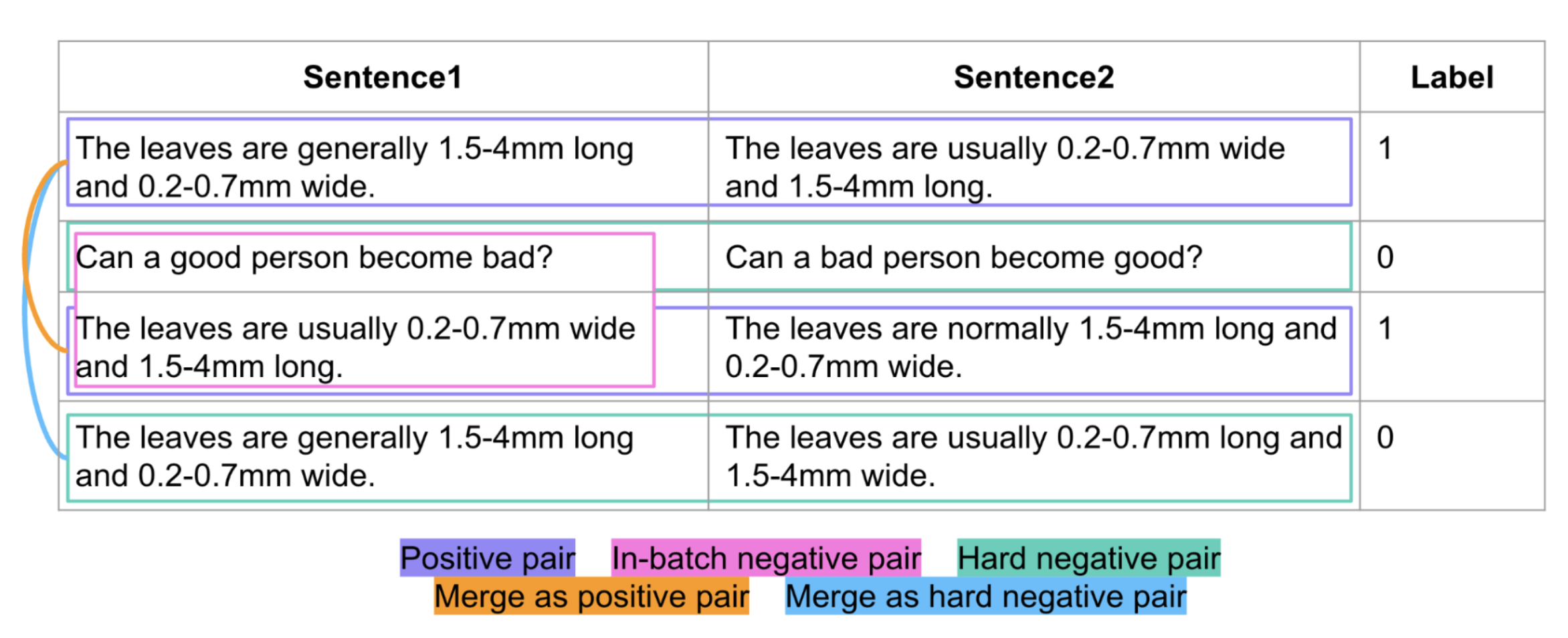}
    \caption{Sampling details}
    \label{fig:enter-label}
\end{figure}
\vspace{\extraspace}
\subsection{Model}

As the base bi-encoder model we decided to use LaBSE \cite{feng2022languageagnostic}. The LaBSE model is designed to learn language-agnostic sentence embeddings. It combines the best methods for learning both monolingual and cross-lingual representations. Key features include:
\begin{itemize}
\item \textbf{Multilingual Capability}: LaBSE supports 109 languages and achieves state-of-the-art performance on various bi-text retrieval and mining tasks. It significantly reduces the amount of parallel training data needed for good performance by 80%.
\item \textbf{Embedding Quality}: LaBSE’s sentence embeddings are language-agnostic, making them suitable for cross-lingual applications. Despite this, they still perform competitively on monolingual transfer learning benchmarks.
\end{itemize}

During training to save model ability to zero-shot transfer to other languages we freeze embedding layer. 
\vspace{\extraspace}
\subsection{Objective}

As a baseline objective, we used the objective from the LaBSE paper. However, continuous fine-tuning with it is not very effective. To improve the training process, we decided to modify the original Additive Margin Softmax Loss by incorporating some ArcFace features \cite{feng2022languageagnostic}. Here’s how we proceed:

\textbf{Creating Vectors:}
\begin{itemize}
\item We create a vector of anchor pairs, denoted as \(x\), and its paraphrase vector, denoted as \(y\). Elements at the same position \(i\) in these vectors are considered similar, while elements at different positions represent in-batch negatives.
\item Additionally, we create a vector \(h\), where we include hard negative examples from the dataset and augment them with hard negative examples from the mega batch.
\end{itemize}

\textbf{Components:}
\begin{itemize}
\item \(\varphi{(x_i, y_i)}\) represents the distance function.
\item \(m\) is the margin.
\item \(s\) and \(g\) are scales used to correct the impact of each component in the loss function.
\end{itemize}

Let’s define the positive component as:
\begin{equation}
p = e^{\text{s*cos}(\varphi{(x_i, y_i)}+m)}
\end{equation}

The negative component is computed as the sum over all in-batch negative examples:

\begin{equation}
n = \sum_{n=1, n\neq i}^{N} e^{\text{s*cos}(\varphi{(x_i, y_n)})}
\end{equation}

We consider hard negative examples from both the dataset and the mega batch: 
\begin{equation}
h = \sum_{j=1, j\neq i}^{HN}  e^{\text{s*cos}(\varphi{(x_i, j_h)})}
\end{equation}

After applying substitutions, we obtain the following objective function \textbf{(Additive Margin Scale Loss)}:
\begin{equation}
L = -\frac{1}{N} \sum_{i=1}^{N} \log \left( \frac{p}{p + n + g*h} \right)
\end{equation}
\vspace{\extraspace}
\subsection{Hard-negative mining}

Using only in-batch negatives is not the most effective or complex way of model training. By limiting the learning procedure in this manner, we encounter an issue: if all potential negative examples are dissimilar from the positive pair, the objective becomes too easy to minimize. To address this, the procedure called ‘mega-batching’ was proposed in \cite{wieting2018paranmt50m}. The original algorithm is described in Algorithm 1.

\begin{algorithm}[h]
\caption{Mega-batching}
\begin{algorithmic}
\STATE Initialize dataloader with set of positive pairs from dataset.   
\WHILE{Training step}
  \STATE Aggregate M mini-batches to create one mega-batch
  \vspace{\extraspace}
  \STATE Select N closest negative examples for each positive
  \vspace{\extraspace}
  \STATE Split back up into M mini-batches.
\ENDWHILE
\end{algorithmic}
\label{alrc_algorithm}
\end{algorithm}

We decided to make mega-batching more complex and select the most challenging examples for the model. We can measure example complexity through the similarity score that the model assigns to positive and negative combinations. Our simple modification is described in Algorithm 2.

\begin{algorithm}[h]
\caption{Proposed hard negative mining algorithm}
\begin{algorithmic}
\STATE Initialize dataloader with set of positive, negative and hard negative from dataset, where hard negative is a negative example from PAWS-X and negative is an example had chosen from batch.   
\WHILE{Training step}
  \STATE Aggregate M mini-batches to create one mega-batch
  \vspace{\extraspace}
  \STATE Select all negative examples that are closer to anchor than some threshold \OR Select N negative examples from mega-batch (the N closest negative examples for each pair)
  \vspace{\extraspace}
  \STATE Split back up into M mini-batches.
\ENDWHILE
\end{algorithmic}
\label{alrc_algorithm}
\end{algorithm}

\subsection{Training setup}
We trained the LaBSE model for 50 epochs on the PAWS-X training data using the Additive Margin Scale Loss with parameters: margin=0.5, scale=0.5, and gamma=1. The mega batch size was set to 20. Our training included five languages from PAWS-X, excluding French and Korean, to assess the model’s zero-shot ability.

%%%%%%%%%%%%%%%%%%%%%%%%%%%%%%%%%%%%%%%%%%%%%%%%%%%%%%%%%%%%%%%%%%%%%%%%%%%%%%%%
\section{Evaluation}
\vspace{\extraspace}

\subsection{Classification strategy}
To classify phrases using cosine similarity of sentence embeddings, we can use different threshold-choosing strategies. We can choose a threshold to maximize metrics such as \textit{accuracy, EER, and F1}. Since the dataset is balanced and our target is accuracy, we use the ‘Max accuracy’ strategy. This strategy means that we choose a threshold that allows us to achieve the best accuracy on the dev set.
\vspace{\extraspace}

\subsection{Align and uniform losses}
For additional evaluation of embedding space quality, we utilized the Align and Uniform losses from SimCSE \cite{Deng_2022}. These losses assess two main characteristics of a ‘good’ embedding space:
\begin{itemize}
\item \textbf{Align}: Measures the expected distance between embeddings of paired instances.
\item \textbf{Uniformity}: Evaluates how uniformly the embeddings are distributed.
\end{itemize}

\vspace{\extraspace}

\subsection{Results}
The evaluation results on PAWS-X test set are presented in Table 1 and Table 2. 

Cross-Encoder and Encoder-Decoder models dominate the field. The main reason for this is that these models process two sentences simultaneously, unlike bi-encoders, which process them independently. Although the gap in performance is not large, the Bi-Encoder model, despite having more parameters, is significantly faster (or conversely: the price for quality is complexity). This speed advantage arises because the Bi-Encoder does not suffer from quadratic complexity and allows for offline computation.

\begin{table}
    \centering
    \begin{tabular}{|c|c|c|c|}
     \hline
         Model&  Params &  Mean accuracy& Model type\\
        \hline
         ByT5 XXL& 2.9B&  91.7& Encoder-decoder\\
         mT5 XXL&  2.9B&  91.5& Encoder-decoder)\\
         ByT5 Small&  300M&  88.6& Encoder-decoder\\
         mT5 Small&  300M&  87.7& Encoder-decoder\\
         mBERT &  177M&  87.2& Cross-encoder\\
         Coupled &  177M&  85.3& Cross-encoder\\
         Decoupled &  177M&  85.0& Cross-encoder\\
         \textbf{LaBSE} & \textbf{471M}& \textbf{79.3}& \textbf{Bi-encoder}\\
          \hline
    \end{tabular}
    \caption{Evaluation results on PAWS-X test set}
    \label{tab:my_label}
\end{table}

\begin{table}
    \centering
    \begin{tabular}{|c|c|c|c|}
     \hline
         Model& Inter-para accuracy& Inter-same accuracy& Intra accuracy\\
        \hline
         LaBSE & 77.4&  82.2& 78.3\\
          \hline
    \end{tabular}
    \caption{Evaluation results on PAWS-X test set divided by paraphrase classes}
    \label{tab:my_label}
\end{table}
%%%%%%%%%%%%%%%%%%%%%%%%%%%%%%%%%%%%%%%%%%%%%%%%%%%%%%%%%%%%%%%%%%%%%%%%%%%%%%%%
\section{Conclusion}

In our research, we made strides in training a bi-encoder for semantic similarity tasks. Specifically, we identified an optimal set of modifications for the Additive Margin Softmax Loss, which improved the quality of paraphrase identification. Additionally, we applied various semantic similarity training techniques. As a result, our model achieved performance comparable to state-of-the-art cross-encoders, with only a minimal relative drop of 7-10\%. These findings highlight the effectiveness of our approach in enhancing semantic similarity modeling. 

\bibliographystyle{ieeetr}
\bibliography{bibliography}

\clearpage

\end{document}